%% file: main.tex
\documentclass[11pt]{article}

\usepackage[final]{acl}

\usepackage{times}
\usepackage{latexsym}

\usepackage[T1]{fontenc}

\usepackage[utf8]{inputenc}

\usepackage{microtype}

\usepackage{inconsolata}

\usepackage{graphicx}
\usepackage{booktabs}
\usepackage{array}
\usepackage{arydshln}
\usepackage{enumerate}

\usepackage{tcolorbox}
\usepackage{paralist}
\usepackage[inline]{enumitem}
\usepackage[symbol]{footmisc}

\title{Neural Models and Language Model Prompting for the Multidimensional Evaluation of Open-Ended Conversations}

\author{
  \textbf{Michelle Elizabeth\textsuperscript{*,1,2}},
  \textbf{Alicja Kasicka\textsuperscript{*, 1}},
  \textbf{Natalia Krawczyk\textsuperscript{*, 1}},\\
  \textbf{Magalie Ochs\textsuperscript{2}},
  \textbf{Gw\'enol\'e Lecorv\'e\textsuperscript{1}},
  \textbf{Justyna Gromada\textsuperscript{1}},
  \textbf{Lina M. Rojas-Barahona\textsuperscript{1}}
\\
  \textsuperscript{1}Orange Research
  \textsuperscript{2}Aix-Marseille University
\\
  \small{\href{mailto:m@domain}{michelle.elizabeth@orange.com},}
  \small{\href{mailto:m@domain}{alicja.kasicka@orange.com},}
  \small{\href{mailto:m@domain}{natalia1.krawczyk@orange.com},}
  \\
  \small{\href{mailto:m@domain}{magalie.ochs@lis-lab.fr},} 
  \small{\href{mailto:m@domain}{gwenole.lecorve@orange.com},}
  \small{\href{mailto:m@domain}{justyna.gromada@orange.com},}
  \small{\href{mailto:m@domain}{lina.rojas@orange.com}}
}

\renewcommand{\thefootnote}{\arabic{footnote}}

\begin{document}

\maketitle


\begin{abstract}

The growing number of generative AI-based dialogue systems has made their evaluation a crucial challenge. This paper presents our contribution to this important problem through the Dialogue System Technology Challenge (DSTC-12, Track 1), where we developed models to predict dialogue-level, dimension-specific scores. 
Given the constraint of using relatively small models (i.e. fewer than 13 billion parameters) our work follows two main strategies: employing Language Models (LMs) as evaluators through prompting, and training encoder-based classification and regression models.
Our results show that while LM prompting achieves only modest correlations with human judgments, it still ranks second on the test set, outperformed only by the baseline.
The regression and classification models, with significantly fewer parameters, demonstrate high correlation for some dimensions on the validation set. Although their performance decreases on the test set, it is important to note that the test set contains annotations with significantly different score ranges for some of the dimensions with respect to the train and validation sets.

\end{abstract}

\def\thefootnote{*}\footnotetext{Equal contribution.}\def\thefootnote{\arabic{footnote}}

\section{Introduction}
Real-life dialogues are unpredictable and dynamic, making them difficult to reproduce in static corpora. Consequently, dialogue systems are typically evaluated with either simulated users or real users~\cite{zhu2022convlab3}. However, a significant gap exists between these approaches, leading to unrealistic simulations or subjective human evaluations~\cite{cordier-etal-2023-shot,elizabeth2025tod}. Despite its subjectivity, human evaluation is preferred.
In the seminal framework PARADISE \cite{paradise1997}, subjective metrics, such as user satisfaction, were estimated based on objective metrics through linear regression.  Reinforcement Learning from Human Feedback (RLHF)~\cite{Christiano2017DeepRL, Ibarz2018RewardLF} utilizes regression models as reward models to evaluate the output of Language Models (LMs)~\cite{10.5555/3600270.3602281} for better alignment to human preferences. This may provide a rationale for high correlation between LM and human judgments~\cite{kazi2024large, gunasekara2020overview}. These results suggest that regression models can be a promising approach to conversation evaluation.

Track~1 of \textsc{DSTC-12} ``Dialog System Evaluation: Dimensionality, Language, Culture and Safety''~\cite{mendonca2025dstc12t1} focuses on automatic evaluation of open-domain dialogues for ten dimensions, at the dialogue level. 
The challenge incorporates widely-used dimensions such as overall quality, \textit{Relevance}, and \textit{Proactivity}, alongside less conventional ones including \textit{Empathy}, \textit{Trust}, and \textit{Skill}. 
This provides a valuable opportunity to assess the correlations between human judgments and automatic evaluation techniques across each dimension.
For this purpose, we present four distinct approaches for dialogue-level evaluation that were submitted to the challenge by our team \textit{ORALIS}.

This work covers three possible representations of the scores and one combination of these approaches:
\begin{enumerate}[label=\roman*]
\item the most straightforward approach, treating scores as real numbers and predicting them through a \textit{regression} task; 
\item treating scores as classes, since they are integers that correspond to categories of evaluation (such as good, average, poor), which leads to training \textit{classifiers}; 
\item treating scores as tokens among others as handled in autoregressive LMs, and thus using \textit{LM prompting} to generate scores. 
\item a final strategy, referred to as the \textit{hybrid} approach, consists of mixing predictions from diverse approaches for various dimensions.
\end{enumerate}

According to the results, none of our systems outperforms the baseline (a prompted \textit{Llama-3.1-8B-Instruct}\footnote{\url{https://huggingface.co/meta-llama/Llama-3.1-8B-Instruct}} LM) in terms of average \textit{absolute} correlation on the test set. However, our approaches outperform the baseline on most individual dimensions. The LM prompting system shows better generalization to the test set than other approaches, performing better on some dimensions than on the validation set.
The regression and classification models demonstrate strong \textit{positive} correlations with human scores on the validation set but achieve lower \textit{absolute} correlation scores when applied to unseen examples, suggesting overfitting.
The classification approach, while ranking lowest overall alongside the hybrid method, excels on six dimensions including \textit{Empathy}, outperforming all other approaches in terms of number of winning dimensions. 
The hybrid method, which selects the best-performing approaches on the validation set (combining LM prompting and regression while excluding classification), does not generalize well to the test data. These results can also be explained by the fact that there are inconsistencies between the training-validation sets and the test set, especially regarding the score distribution and score ranges as depicted in  Figure~\ref{fig:h_annot} and Figure~\ref{fig:test_h_annot}.

The paper is organized as follows: Section~\ref{sec:related_work} provides a literature review on dialogue evaluation; Section~\ref{sec:dataset_and_dimensions} introduces the datasets and dimensions used in our work, while Section~\ref{sec:evaluators} details the four implemented evaluators. Finally, Section~\ref{sec:results} reports the results of the validation set (as used to develop the evaluators) as well as on the test set (as used to rank the submitted evaluators in the challenge).

\section{Related Work}
\label{sec:related_work}

This section discusses evaluation paradigms, recent advances in automatic and LM-based metrics, current multi-dimensional frameworks and open challenges.

Open-ended conversational AI systems require multi-dimensional assessment due to the complex nature of dialogue, where multiple valid responses exist for any given context. Key dimensions include \textit{coherence} (the contextual appropriateness and logical consistency of responses~\cite{Baoetal2021}), \textit{engagement} (sustaining user interest~\cite{Venkateshetal2018}), \textit{informativeness} (providing relevant content~\cite{Baoetal2021}), \textit{specificity} (context-tailored responses~\cite{Harrisonetal2023}), \textit{consistency} (avoiding contradictions~\cite{Baoetal2021}), and \textit{factual correctness} (minimizing hallucinations and ensuring accurate information~\cite{Baoetal2021}). Additional dimensions include \textit{fluency}, \textit{personality}, and \textit{context management} (maintaining memory across multiple turns~\cite{Wangetal2024}).

Traditional reference-based metrics, e.g. BLEU or ROUGE, show weak correlation with human judgments in open-domain dialogue~\cite{Liuetal2016,Salehetal2020}. While human evaluation remains most reliable~\cite{Lietal2019,Venkateshetal2018}, it is costly, time-consuming, and can suffer from inconsistency~\cite{Jietal2022, Smithetal2022}.

Researchers have developed various automatic metrics to overcome evaluation challenges. Embedding-based metrics capture semantic similarity beyond surface-level lexical overlap but struggle with catching conversational nuances. Regression models, inspired by PARADISE~\cite{paradise1997} and RLHF~\cite{Christiano2017DeepRL}, and classification models are constrained by availability and quality of training data. Reference-free metrics like FED~\cite{Mehrietal2020} evaluate responses in context with better human alignment. LM-based evaluation uses LMs as judges, showing stronger correlation with human ratings~\cite{Linetal2023,Yuetal2024}, despite challenges including self-preference bias~\cite{chen2025llm} and sensitivity to response characteristics and context complexity~\cite{xu2025does}. 

The adoption of LMs as judges~\cite{gunasekara2020overview,kazi2024large} enables scalable evaluation, although concerns about annotation quality persist. Small LMs offer cost-effective alternatives and have recently shown strong potential as capable judges. Although they may seem less accurate due to their size, recent work~\cite{deshpande2024glider} shows that well-aligned LMs with around 3B parameters can achieve the performance of much larger systems.

Evaluation campaigns like DSTC-9~\cite{gunasekara2020overview} and DSTC-11~\cite{soltau-etal-2023-dstc} have advanced the field through interactive and multilingual evaluation tracks. However, the most successful systems still rely heavily on fine-tuned generative models or LMs for data augmentation, and achieving high correlation with human judgments remains a challenge, especially for multi-dimensional conversation aspects.

Modern evaluation frameworks assess conversation quality across several interdependent dimensions like coherence, engagement, and context management~\cite{Baoetal2021,Harrisonetal2023, Wangetal2024}. Multi-dimensional LM-based approaches (e.g. LLM-Eval~\cite{Linetal2023}, MT-Bench~\cite{Baietal2024}, KIEval~\cite{Yuetal2024}) offer thorough assessments yet face challenges with bias, generalizability, and scalability.

Despite significant progress in automatic evaluation of conversational systems, current methods still face limitations in robustness, interpretability, and scalability, highlighting the need for improved multi-dimensional approaches that reliably reflect human perceptions of conversational quality across diverse scenarios.

\section{Datasets and Dimensions}
\label{sec:dataset_and_dimensions}

We use three datasets in this work: \textsc{DSTC-12} ~\cite{mendonca2025dstc12t1}, the official competition dataset; \textsc{ConTurE}~\cite{gunasekara2020overview}, which was used in the DSTC-9 evaluation campaign and FED~\cite{mehri-eskenazi-2020-unsupervised}, another dataset published for dialogue evaluation research.

As detailed later in this section, these datasets predominantly contain open-ended human-machine dialogues annotated by humans on dialogue-level for various evaluation metrics, although the \textsc{FED} dataset also includes human-human dialogues.

\label{s:data}
\subsection{\textsc{DSTC-12} Dataset and Metrics}
\paragraph{\textsc{DSTC-12}} ~\cite{mendonca2025dstc12t1} is an official dataset released as part of the competition. It contains $185$ open-domain human-machine dialogues in English. Each dialogue covers a wide variety of everyday topics such as personal stories, preferences and recommendations, and fact-based planning queries. Detailed dataset statistics are displayed in Table \ref{table:data_stats}. What is worth noting is the significant difference between the length of utterances in \textsc{DSTC-12}, compared to other datasets.
Still, in all datasets, the average number of words in machine turns is significantly greater than in human utterances.

The dialogues were evaluated by human annotators across ten dimensions. 
According to the organizers, human annotators were either MTurk workers or lab members, thus different dialogues might have been annotated by different annotators.
This combination of research staff and online workers might raise concerns about potential inconsistencies in how dimensions were scored across conversations.

Unfortunately, the annotated data is highly imbalanced, as not all dialogues have scores assigned for each evaluation dimension.

\begin{figure}[ht!]
    \centering
    \includegraphics[width=\linewidth]{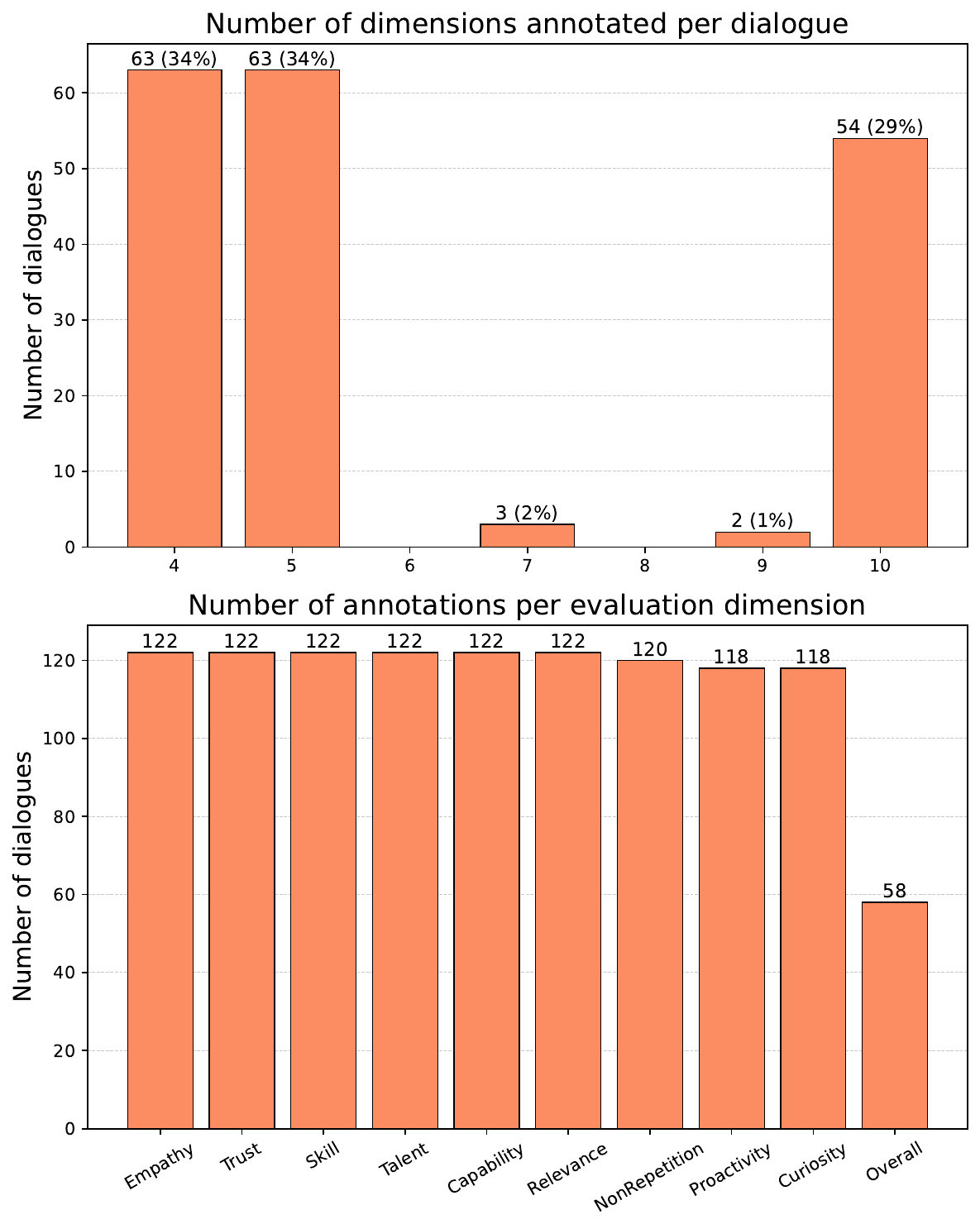}
    \caption{Distribution of the dialogues in the DSTC-12 dataset (train/validation) based on the scores (top), and the dimension (bottom).}
    \label{fig:ann_coverage}
\end{figure}

While each dialogue received scores for at least four dimensions, the coverage varies considerably. Only 54 out of 185 dialogues (29\%) are annotated with all ten dimensions, while the majority include annotations for only four or five. At the dimension level, annotation counts are also unevenly distributed: most dimensions are well represented with over 120 annotations, whereas \textit{Overall} appears in less than one-third of the dialogues. These patterns are illustrated in Figure~\ref{fig:ann_coverage}. The test set consists of $120$ dialogues, with varying coverage of annotations as in the training set.

In addition, dimensions have different score ranges. Their names, along with their ranges, are:  \textit{Empathy} (1-12), \textit{Trust} (0-5), \textit{Curiosity} (0-100), \textit{Proactivity} (0-100), \textit{NonRepetition} (0-100), \textit{Relevance} (0-100), \textit{Overall} (0-100), \textit{Skill} (0-5), \textit{Talent} (0-5), and \textit{Capability} (0-5).
For dimensions such as \textit{Relevance}, \textit{NonRepetition}, \textit{Proactivity}, and \textit{Curiosity}, the majority of the human scores are between 6 and 10, despite the score range being 0-100. The distributions of the scores are uneven, particularly for dimensions with the 0-100 range, see Appendix \ref{appendix:additional_figures}, Figure \ref{fig:h_annot}. 
In the test set, the score ranges are between 1-5 for \textit{Skill, Talent, Capability, Trust, and Overall} while the range is 1-10 for \textit{Empathy, Relevance, NonRepetition, Proactivity and Curiosity}, which differs from the score ranges observed in the training set. See Appendix \ref{appendix:additional_figures}, Figure \ref{fig:test_h_annot}.

\begin{table}[]
    \centering
    \scalebox{0.8}{
    \begin{tabular}{l|c c|c|c}
        & \multicolumn{2}{c|}{DSTC-12} & FED & ConTurE  \\
        & train & test& & \\
        \hline
         \#Dialogues & 185 & 120 & 125 & 119  \\
         \#Ann. per Dialogue & 1 & 1 &  5 & 3  \\
         Avg. \#turns & 15 & 21 & 6 & 9  \\
         Avg. \#words per turn (H) & 25 & 51 & 6 & 7  \\
         Avg. \#words per turn (M) & 130 & 193 & 12 & 19  \\
         \hline
    \end{tabular}}
    \caption{Statistics for all the datasets (Ann. stands for annotations, H stands for human and M stands for machine).}
    \label{table:data_stats}
\end{table}
\paragraph{Metrics: }
The challenge assesses the evaluators based on the mean \textit{absolute} Spearman correlation with human judgments. In our experiments, we also consider the mean \textit{positive} correlation with human judgments.

\subsection{FED and ConTurE}
\label{ss:extdata}
The official dataset released for the challenge, i.e. \textsc{DSTC-12}, is rather small, containing only $185$ dialogues. Each dialogue was annotated by only one evaluator. 
To increase data diversity and avoid overfitting when training our regression and classification models, we utilized two other open-domain human–machine dialogue datasets: \textsc{ConTurE}~\cite{gunasekara2020overview}, with $119$ dialogues, which was proposed in DSTC-9 Track 3, and \textsc{FED}~\cite{mehri-eskenazi-2020-unsupervised}, with $125$ dialogues, introduced at SigDial 2020. 
In all these three datasets, the dialogues predominantly involve interactions between a human and a machine, although the \textsc{FED} dataset also includes human-human interactions with one participant simulating a machine. The conversations cover a variety of everyday topics such as personal preferences, opinions, popular culture, and general knowledge, resembling natural and informal interactions.
 
In contrast to \textsc{DSTC-12} dataset, the \textsc{ConTurE} dataset has each dialogue annotated by three different raters, while in \textsc{FED} dataset each dialogue was annotated by five different evaluators.
To ensure that every dialogue contributes exactly one score per dimension to our models, just as in the \textsc{DSTC-12} dataset, and to prevent over-representation of multi-rated dialogues, we first averaged the dimension scores in both \textsc{ConTurE} and \textsc{FED}. 
Then, we mapped the dimension names from the additional datasets to match those in the \textsc{DSTC-12} dataset. This mapping was based on the heuristics shown in Table~\ref{t:mapping}, where we paired each metric from \textsc{ConTurE} and \textsc{FED} with the \textsc{DSTC-12} dimension that best reflected its core intent.

\begin{table}[]
    \centering
    \scalebox{0.85}{
    \begin{tabular}{ccc}
         \textsc{FED} \& \textsc{ConTurE} &  & \textsc{DSTC-12}  \\\hline
         Inquisitive & ---> & Curiosity\\
         Avg(Informative, Coherence) & ---> & Relevance \\
         Topic depth & ---> & Talent \\ 
         Flexible & ---> & Proactivity\\ 
         Diverse & ---> & Non-repetition\\
         Likeable & ---> & Empathy \\
         Consistent & ---> & Trust \\ 
         Understanding & ---> & Capability\\ 
         Error recovery & ---> & Skill   \\\hline      
    \end{tabular}}
    \caption{ \textsc{FED} \& \textsc{ConTurE}  to \textsc{DSTC-12} mapping.}
    \label{t:mapping}
\end{table}

In both external datasets, the dimensions are annotated on different scales: for most dimensions, annotations are on a 3-point scale, except for Consistent, which is binary, and Overall quality, which is on a 5-point scale.

After averaging the raw scores per dialogue, we applied a linear rescaling function to fit each dimension into the corresponding dimension range in the \textsc{DSTC-12} dataset:

\begin{equation}
 q = (p - A)\,\frac{D - C}{B - A} + C   
 \label{eq:1}
\end{equation}

where \(p\) is the original value in \([A,B]\) and \(q\) is the mapped value in \([C,D]\). Finally, we rounded \(q\) to the nearest integer to obtain the final score on the target dimension scale in the \textsc{DSTC-12} dataset. 
We split the official DSTC-12 dataset equally into train and validation sets for each dimension. The training set was further enhanced by adding the dialogues from the two additional datasets.

\section{Evaluators}
\label{sec:evaluators}

In this section, we first describe the baseline system (provided by the challenge organizers) and then present the evaluation systems submitted to the challenge.
Our first approach is based on the LM-as-a-judge paradigm, namely \textit{LM Prompting}. The next two systems are classic neural models: \textit{regression} and \textit{classification}. 
Finally, the last system is a hybrid evaluator that combines the \textit{regression} model with the \textit{LM Prompting} system.

\subsection{Baseline} This system was proposed by the organizers and it is a fine-tuned \textit{Llama-3.1-8B-Instruct}\footnote{\url{https://huggingface.co/meta-llama/Llama-3.1-8B-Instruct}} pretrained model for content safety classification, prompted for dialogue-level evaluation. For all of the dimensions the same prompt template was used, which included all evaluation dimensions in one prompt.

\subsection{LM Prompting} 
We tested various prompting methods:
(i) basic prompt with just the name of the dimension, see example in Appendix \ref{appendix:lm-prompts-bp}; (ii) zero-shot learning with the definition of the dimension, see example in Appendix \ref{appendix:lm-prompts-0sp}; (iii) one-shot learning, see example in Appendix \ref{appendix:lm-prompts-1sp}; (iv) few-shot learning (with 3 samples), see example in Appendix \ref{appendix:lm-prompts-fsp} and (v) self-consistency prompting, see example in Appendix \ref{appendix:lm-prompts-scp}.

We assigned various roles to the LM (such as "crowd-worker", "expert", or "human evaluator") and provided task descriptions with score ranges at varying levels of detail.

For one-shot and few-shot learning methods, we provided examples with their assigned scores in two formats: either as conversation excerpts or as summarized dialogue.
For the few-shot learning, we randomly sampled three dialogues:  one with the lowest possible score, one from the median, and one with the highest possible score. For one-shot learning, we randomly sampled a dialogue with a score around the median.
In self-consistency prompting, the LM was provided with a short description of the meaning of the scores within the score range, as well as was asked to check the validity of its response and fix it, if needed, before responding. 

In every prompting method, the LM was asked to return the score along with a short explanation for the given score. Various versions of the prompts were evaluated on each dimension separately, and different prompts were selected for later study based on these preliminary results (listed in the Appendix \ref{appendix:lm-prompts}).

We explored several dialogue context strategies to feed the LM: the last 40\% of the conversation, the first 40\% of the conversation, the first 20\% and last 20\% of the conversation, a summarized version of the dialogue, or the full dialogue.
The summarised version was obtained by summarising each utterance using \textit{Llama 3.1 8B Instruct} model, as it performs well across variety of tasks \cite{grattafiori2024llama}. 
We tested two different summarisation prompts for sentences whose length exceeded 200 words, in conversations with more than $3000$ words in total. These prompts mainly differed by the maximum number of tokens in their summarised versions: either $50$ (\textit{summarisation 1}) or $150$ (\textit{summarisation 2}). The exemplary prompt can be found in the Appendix, section \ref{lm-prompts-summ}.

We considered three state-of-the-art LMs: \textit{Deepseek Llama 8B}\footnote{\url{https://huggingface.co/deepseek-ai/DeepSeek-R1-Distill-Llama-8B}}, \textit{Deepseek Qwen 7B}\footnote{\url{https://huggingface.co/deepseek-ai/DeepSeek-R1-Distill-Qwen-7B}}, and \textit{Qwen 2.5 7B Instruct 1M}\footnote{\url{https://huggingface.co/Qwen/Qwen2.5-7B-Instruct-1M}}. These models were selected based on their demonstrated effectiveness on multiple natural language processing tasks \cite{guo2025deepseek, yang2025qwen2}.

\begin{table*}[!h]
\centering 
\begin{tabular}{| l | l | l | l |} 
\hline 
Dimension & Prompting & Dialogue part & Language Model \\ [0.5ex] 
\hline\hline
Empathy & zero-shot & last 40\% & Qwen 2.5 7B Instruct \\ 
\hline
Trust & zero-shot & full conversation & Qwen 2.5 7B Instruct \\
\hline
Skill & zero-shot & first 20\% + last 20\% & Qwen 2.5 7B Instruct \\
\hline
Talent & zero-shot & summarisation 1 & Qwen 2.5 7B Instruct \\
\hline
Capability & zero-shot & summarisation 1 & Qwen 2.5 7B Instruct \\
\hline
*Capability & zero-shot & summarisation 2 & Deepseek Qwen 7B \\
\hline
Relevance & few-shot & summarisation 1 & Deepseek Llama 8B \\
\hline
*Relevance & few-shot & first 20\% + last 20\% & Qwen 2.5 7B Instruct \\
\hline
NonRepetition & few-shot & first 40\% & Deepseek Qwen 7B \\
\hline
*NonRepetition & few-shot & first 20\% + last 20\% & Qwen 2.5 7B Instruct \\
\hline
Proactivity & zero-shot & first 40\% & Deepseek Llama 8B \\
\hline
*Proactivity & zero-shot & first 20\% + last 20\% & Qwen 2.5 7B Instruct \\
\hline
Curiosity & zero-shot & summarisation 2 & Deepseek Qwen 7B \\
\hline
*Curiosity & zero-shot & first 40\% & Deepseek Llama 8B \\
\hline
Overall & zero-shot & full conversation & Qwen 2.5 7B Instruct \\ [1ex]
\hline 
\end{tabular}
\caption{LM prompting approach: Chosen methods and models for each evaluation dimension, * refers to the combination that obtained the highest absolute correlation on the validation dataset.} 
\label{table:lm_det} 
\end{table*}

We tested various combinations of prompting on the validation set. We modified the prompt, the dialogue context, and utilized distinct LMs. Then, we selected the best performing combination for each dimension, i.e., achieving the highest \textit{positive} correlation values with human annotations. The chosen configuration for each dimension is shown in Table \ref{table:lm_det}. 

Analysis of the standard deviation values for correlation results for different models (average std=0.09) and for different dialogue contexts (average std=0.12) implies that the latter, on average, impacts the final correlation result slightly more than the choice of the model. 

Systems' performances on the validation set and test set are presented in Table \ref{table:corr_val} and Table \ref{table:corr_test}, respectively.

\subsection{Regression}
We trained a regression model for each dimension. The model's architecture consists of a regression layer on top of a ModernBERT Large encoder~\cite{modernbert}. It is worth noting that ModernBERT has a context limit of 8K tokens allowing for encoding a larger dialogue context, in contrast to BERT-family models~\cite{devlin-etal-2019-bert}, which are limited to only $512$ tokens. ModernBERT is also notably smaller than LMs, with fewer than 1 billion parameters (395 million).
We utilized the score ranges provided in the challenge dataset and the mean-square error as the loss function. 
To generalize better and avoid overfitting, we utilized \textsc{ConTurE} and \textsc{FED} datasets in addition to the \textsc{DSTC-12} dataset, using the mapping introduced in Section~\ref{ss:extdata}.
Regression performance on the validation and test sets is presented in Table~\ref{table:corr_val} and Table~\ref{table:corr_test}, respectively.
A later experiment with varying values of weight decay for training, did not show any considerable improvement in the correlations on the test set.

\subsection{Classification}

Similar to the regression system, we trained individual classifiers for each dimension on our combined training set. All dialogues were encoded using \textit{Sentence-BERT (SBERT)}\footnote{\href{https://huggingface.co/sentence-transformers/all-MiniLM-L6-v2}{https://huggingface.co/sentence-transformers/all-MiniLM-L6-v2}}. Since we require discrete categories, each integer score was rescaled to an integer range using Equation~\ref{eq:1} and rounded to the nearest integer.

Model development followed a two-stage grid search. In the first stage, we explored different class ranges and selected [0, 8] based on validation performance. In the second stage, we tuned  \textit{Multi-Layer Perceptron (MLP)} hyperparameters separately for each dimension to maximize \textit{positive} validation Spearman correlation. To reduce overfitting, we specifically optimized regularization and training duration balancing convergence with generalization. The best hyperparameters for each dimension were used to train the final classifiers and predict dimension scores on the test set. 
Predicted scores were rescaled back into the original ranges, using the inverse of Equation \ref{eq:1}.
The validation and test performance of our classifiers trained with SBERT encodings are presented in Table~\ref{table:corr_val} and Table~\ref{table:corr_test}, respectively.

To further address potential overfitting, we applied ModernBERT encodings as in the regression models, combined with label smoothing. However, these changes resulted in slightly lower test set correlations, suggesting that increased model capacity was not sufficient to improve performance.

\begin{table*}[ht]
\centering 
\begin{tabular}{| l | c | c | c | c |} 
\hline 
Dimension & LM prompting & Regression & Classification & Hybrid \\ [0.5ex] 
\hline\hline
Empathy & 0.3  & 0.23 & \textbf{0.35} & 0.3 \\ 
\hline
Trust & \textbf{0.38} & -0.02 &  -0.07 & 0.38  \\
\hline
Skill & \textbf{0.33} & -0.09 & -0.06 &  0.33  \\
\hline
Talent & 0.26 & \textbf{0.41} &  0.15 & \textbf{0.41 }  \\
\hline
Capability & 0.17 & \textbf{-0.21} &0.00 &  0.17 \\
\hline
Relevance & 0.19  &  \textbf{0.79} & 0.71 & 0.79  \\
\hline
NonRepetition & 0.16 & \textbf{0.75} & 0.68 & 0.75  \\
\hline
Proactivity & 0.01 &\textbf{0.79}  & 0.66 &  0.79  \\
\hline
Curiosity & -0.02 &  \textbf{0.68} & 0.65 & 0.68 \\
\hline
Overall & 0.4 &  0.27 &  \textbf{0.49 }& 0.4 \\
\hline\hline
Abs. Average & 0.22 & \textbf{0.42} & 0.38 &  \textbf{0.5}  \\ [1ex] 
\hline 
\end{tabular}
\centering\caption{Correlation between the gold labels and system's outputs on the validation set for each system. \\
\centering\textbf{Bold} values indicate the highest \textit{absolute} correlation across all systems.} 
\label{table:corr_val} 
\end{table*}

\subsection{Hybrid}
The hybrid system combines methods that performed well on the validation set for each dimension, as shown in Table~\ref{table:corr_val}. When selecting these methods, we limited our choice to only regression and LM prompting approaches, excluding the classification method from consideration. This decision was based on the prioritisation of approaches with stronger contextual understanding and generalization capabilities. In the regression system we utilized ModernBERT that has a larger context window, in comparison to the SBERT model used in the classifier. This allows us to process more dialogue context. We strategically chose LM prompting for several dimensions due to its demonstrated ability to generalize well to unseen data \cite{wang2023large}. Additionally, our classification approach returns discrete integer values, e.g., on the scale 0-8, requiring mapping to, e.g., the 0-100 scale, and potentially introducing approximation errors, while both regression and prompting methods produce continuous values within the desired range without the need for additional mapping. This combination of enhanced contextual processing and a potential for better generalization influenced the choice of methods for our hybrid system. 

It is worth noting that all three approaches, i.e. LM prompting, regression, and classification, were submitted to the challenge separately.

The regression system was chosen for the following dimensions: \textit{Talent}, \textit{Relevance}, \textit{NonRepetition}, \textit{Proactivity}, and \textit{Curiosity}. 
For the remaining dimensions, i.e. \textit{Empathy}, \textit{Trust}, \textit{Skill}, \textit{Overall}, and \textit{Capability}, the LM-prompting was chosen as it obtained the most promising results on the validation set, see Table~\ref{table:corr_val}.
The results of our hybrid system on the test set are shown in Table~\ref{table:corr_test}. This system underperformed on this dataset in comparison to its scores on the validation set. 

\section{Results}
\label{sec:results}

The results of our systems on the test set are presented in Table~\ref{table:corr_test}, along with the baseline approach published by the \textsc{DSTC-12} challenge organizers.

\begin{table*}[!ht]

\centering 
\begin{tabular}{| l | c | c | c | c || c |} 
\hline 
Dimension & LM prompting & Regression & Classification & Hybrid & \textbf{Baseline}\\ [0.5ex] 
\hline\hline
Empathy & -0.08 & \textbf{0.17} & \textbf{-0.17} & -0.08  & 0.06 \\ 
\hline
Trust & 0.01 & \textbf{0.2} & 0.13 & 0.01 & -0.11 \\
\hline
Skill & \textbf{-0.22} & 0.07 & -0.02 & \textbf{-0.22} & -0.1 \\
\hline
Talent & 0.05 & \textbf{0.24} & 0.22 & \textbf{0.24} &  0.1 \\
\hline
Capability & 0.13 & \textbf{0.24} & 0.12 & 0.13 & 0.07 \\
\hline
Relevance & 0.08 & -0.1 & \textbf{-0.28} & -0.1 & 0.23 \\
\hline
NonRepetition & 0.11 & 0.14 & -0.0 & 0.14 & \textbf{0.39} \\
\hline
Proactivity & \textbf{-0.15} & 0.08 & 0.2 & 0.08 & -0.02\\
\hline
Curiosity & \textbf{0.37} & 0.09 & 0.08 & 0.09 & 0.23 \\
\hline
Overall & \textbf{0.31} & 0.13 & -0.17 & \textbf{0.31} & \textbf{0.38} \\
\hline\hline
Abs. Average & \textbf{0.15} & \textbf{0.15} & 0.14 & 0.14 & \textbf{0.17} \\ [1ex] 
\hline 
\end{tabular}
\centering\caption{Correlation between the gold labels and systems' outputs on the test set.\\
\centering\textbf{Bold} values indicate the highest \textit{absolute} correlation across all systems.} 
\label{table:corr_test} 
\end{table*}

The baseline is based on the LM-as-a-judge approach, similar to one of our systems; however, it uses a different LM and different prompt.

The \textit{absolute} average correlation on the test set for all systems is relatively low, between 0.14 and 0.15, while the baseline achieves 0.17. This represents a significant decrease from the validation set, where the regression and hybrid systems achieved values between 0.4 and 0.5 (see Table~\ref{table:corr_val}).

None of our systems achieved a higher average \textit{absolute} score than the baseline; however, our approaches outperform the baseline on most of the individual dimensions. The baseline has higher scores only for the \textit{NonRepetition} and \textit{Overall} dimensions. Nevertheless, the difference on the \textit{NonRepetition} dimension is significant enough to influence the absolute average score for the whole system.

Each of our approaches outperforms the baseline on multiple dimensions in terms of the \textit{absolute} score. The classification approach performs best, in terms of number of winning dimensions, exceeding the baseline on six dimensions, while the LM prompting, regression, and hybrid approaches each outperform on five dimensions. All four of our systems outperform the baseline on \textit{Empathy}, \textit{Capability} and \textit{Proactivity}, and three of them excel on \textit{Talent} as well.

Performance patterns vary across dimensions. The classification approach maintains its strength for \textit{Empathy} from validation to test set in terms of \textit{absolute} correlation, though with reduced values. For \textit{Talent} and \textit{Capability}, the regression system outclasses other approaches across both sets. However, some dimensions show inconsistent results, for example, LM prompting excels on \textit{Trust} on the validation set but its performance drops significantly on the test set. On the test set, the regression system shows the opposite trend for this dimension, performing better than on the validation set.

We observe significant performance decrease between validation and test sets for several dimensions for regression and classification systems, suggesting potential overfitting.  The regression system shows drastic decreases for \textit{Relevance}, \textit{NonRepetition}, \textit{Proactivity}, and \textit{Curiosity}, despite achieving correlations of 0.68-0.79 on the validation set. The classification system demonstrates similar patterns on the same dimensions, with correlations of 0.65-0.71 on the validation set. Nevertheless, it maintains superior performance for \textit{Relevance} on the test set.

Interestingly, LM prompting demonstrates the opposite pattern for some dimensions, performing better on the test set than on the validation set. It achieved the highest \textit{absolute} correlations on test set for \textit{Proactivity}, \textit{Curiosity}, and the \textit{Overall} dimension, despite weaker results on the validation set.

Inspecting both Table~\ref{table:corr_val} and Table~\ref{table:corr_test} raises concerns about why some dimensions show \textit{negative} correlation values.
One possible explanation lies in the conceptual mismatch between how LMs and humans interpret evaluation metrics. The inconsistent score ranges between the training and test set also leads us to question the quality of the annotations. Dimensions may have been understood differently by annotators and models, leading to inconsistent judgments that weakened or even inverted expected correlations. Evaluation systems often reflect individual user experiences shaped by emotion and subjectivity, making consistent human assessment especially difficult~\cite{fan2020survey}. Furthermore, scoring chatbot responses remains a fundamentally subjective and challenging task even for human evaluators, which increases the likelihood of annotation noise in human labels~\cite{yuwono2019automated}.

\section{Conclusions and Future Work}
In this paper, we present four distinct dialogue-level evaluators for different dimensions that were submitted to the \textsc{DSTC-12} challenge.
We explored distinct prompting strategies, including varying the dialogue context across different LMs.
We also trained very small regression and classification models on the challenge dataset enriched with other evaluation datasets (\textsc{ConTurE} and \textsc{FED}).
We also considered a hybrid system that combines the LM prompting and regression approaches. Furthermore, we analyzed the data and found that there are inconsistencies between the training-validation sets and the test set, in terms of the score distribution and score ranges. 
Although our systems did not outperform the baseline, classical approaches, such as regression and classification, show interesting results, competitive with larger models of 7 and 8 billion parameters used in LM prompting approach.

In terms of future work, we first suggest enhancing the quality of the dataset in a dedicated annotation campaign. Second, we would like to explore domain adaptation techniques for training models on similar but larger datasets from distinct sources (such as the \textit{DSTC-11} dataset) to overcome data scarcity. 


\section{Limitations}
The scaling laws have shown that the impressive capabilities of LLMs are highly influenced by three factors: the size of the model, the size of the dataset, and the amount of computing power used for training~\cite{kaplan2020scalinglawsneurallanguage}. 
All LMs used in our experiments have fewer than 13B parameters. The regression and classification models have fewer than 1B parameters.  

We tuned our systems to maximize the \textit{positive} correlation, however the systems were ranked based on the \textit{absolute} correlation.

Moreover, the dataset provided in the challenge is quite small, making it difficult to use for training regression and classification models. The mapping we made between the annotation of the additional datasets and the \textsc{DSTC-12} dataset is entirely subjective, which may require in depth investigation to study the impact of various mappings. It would have been beneficial to have the instructions provided to human annotators for a more accurate mapping as well as to define the dimensions more accurately for the LM prompting method.

Finally, there are concerns regarding the consistency of the annotations for certain dimensions, since their score ranges vary significantly between the train-validation and the test sets.

\section*{Acknowledgments}
This work was granted access to the HPC resources of IDRIS under the allocations AD011015150R1 made by GENCI.
\bibliography{custom}

\input{appendix}

\end{document}

%% file: appendix.tex
\appendix
\section{Additional figures}
\begin{figure*}[ht!]
    \centering
    \includegraphics[width=1\textwidth]{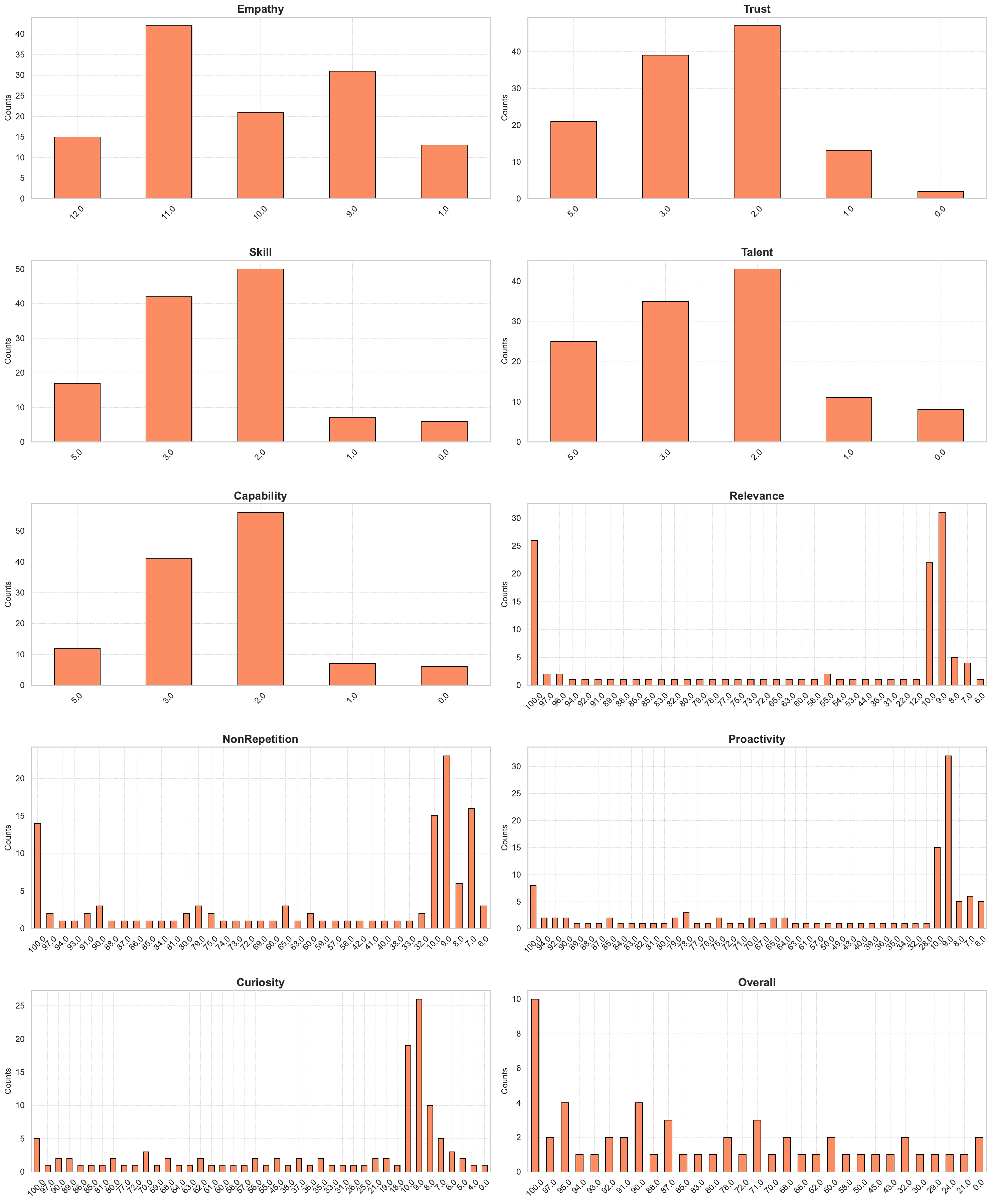}
    \caption{Distribution of human annotations by evaluation dimension in the training set.}
    \label{fig:h_annot}
\end{figure*}

\begin{figure*}[ht!]
    \centering
    \includegraphics[width=1\textwidth]{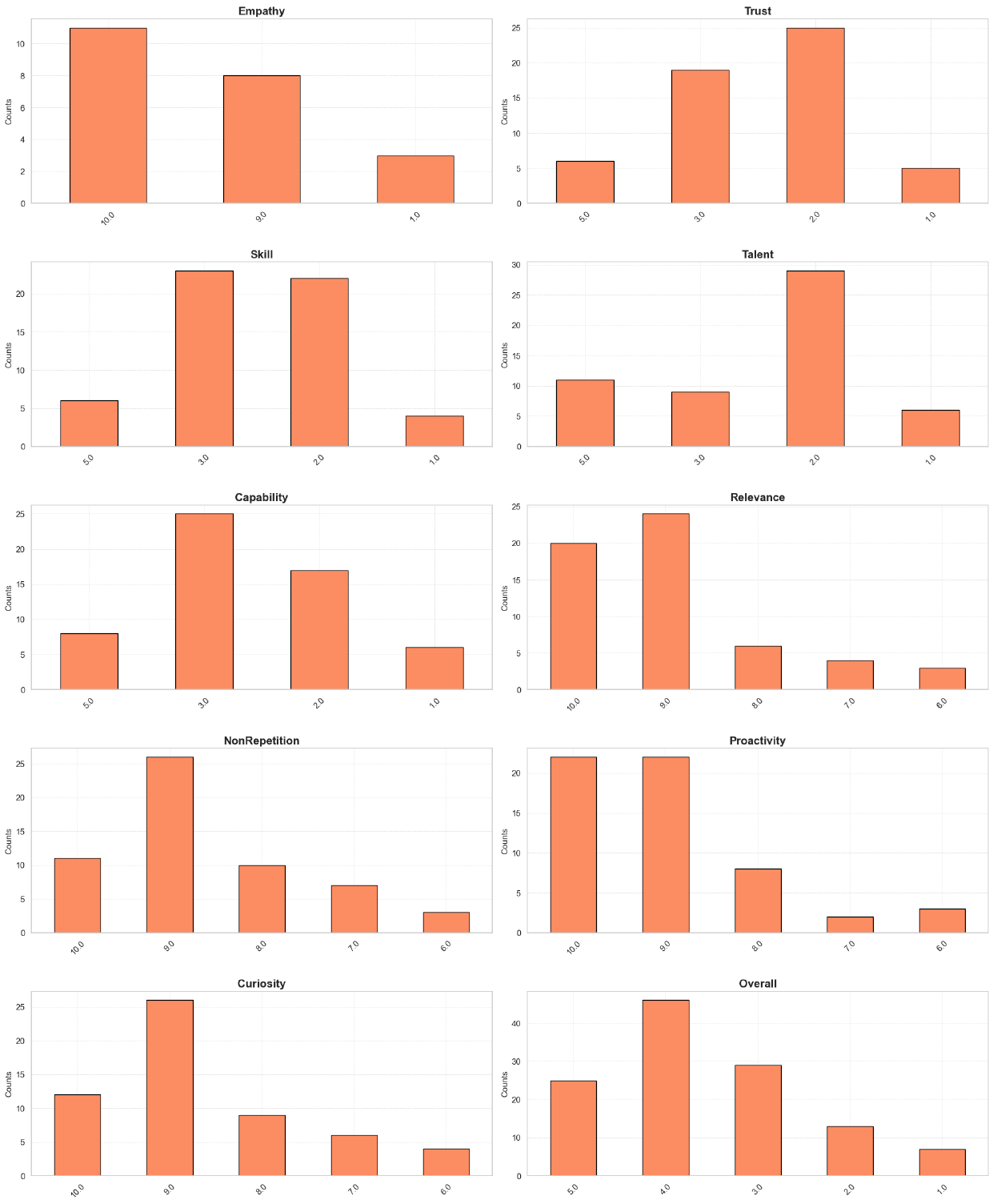}
    \caption{Distribution of human annotations by evaluation dimension in the test set.}
    \label{fig:test_h_annot}
\end{figure*}

We provide the distribution of human annotations by dimension for both datasets provided by the organizers: the training set in Figure~\ref{fig:h_annot}, and test set in Figure~\ref{fig:test_h_annot}.
\label{appendix:additional_figures}

\section{Selected Prompts} \label{appendix:lm-prompts}
In this section we present the selected prompts.
\subsection{Relevance}
{\small \tt You are an expert evaluator tasked with assessing the relevance of chatbot's answers. \\
Relevance refers to the system's ability to provide answers that are related or useful to what is happening or being talked about. \\
Please, evaluate queries of the chatbot in the following conversation by assigning it a score from the scale 0-100, where 0 means that the chatbot's answers are often irrelevant, and 100 suggests that the chatbot's answers are always relevant. \\
The final output should include the score (0-100) and your explanation for the given score. \\
Here are the examples of the excerpts of the conversations and the score these conversations received. Chatbot's and user's utterances are separated using ''';'''. \\
Excerpt from the example conversation: '''\{excerpt1\}''' \\
Score for the example conversation: '''\{score1\}''' \\
(...) \\
The conversation for evaluation: \\
\{conversation\}
}

\subsection{Proactivity}
{\small \tt Act like a human evaluator tasked with assessing the proactivity of chatbot queries. \\
Proactivity refers to the system's ability to anticipate user's future problems, needs, and changes. A proactive chatbot often takes initiative and guides the conversation. \\
Please, evaluate queries of the chatbot in the following conversation by assigning it a score from the scale 0-100, where 0 means that the chatbot is not proactive at all, and 100 suggests that the chatbot often takes initiative and anticipates the needs of the user. \\
The final output should include the score (0-100) and your explanation for the given score. \\
The conversation for evaluation: \\
\{conversation\}
}

\subsection{NonRepetition}
{\small \tt Act like a human evaluator tasked with assessing the chatbot's ability to avoid repeating responses within a conversation. \\
Non-repetition refers to the system's ability to avoid repeating information or questions the user has already provided. A chatbot with strong non-repetition capabilities ensures a smoother conversation by recognising and adapting to previously shared inputs. \\
Please, evaluate queries of the chatbot in the following conversation by assigning it a score from the scale 0-100, where 0 means that the chatbot often repeats itself, and 100 suggests that the chatbot has strong non-repetition capabilities. \\
The final output should include the score (0-100) and your explanation for the given score. \\
Here are the examples of the summaries of the conversations (you will be evaluating a full conversation, not the summary) and the score these conversations received. \\
Summary of the example conversation: '''\{summary1\}''' \\
Score for the example conversation: '''\{score1\}''' \\
(...) \\
The conversation for evaluation: \\
\{conversation\}
}

\subsection{Trust}
{\small \tt You are an expert evaluator tasked with assessing how trustworthy the chatbot seems to the user. Trustworthy chatbot is a chatbot that seems sincere, reliable, and honest, whose responses seem true and not harmful or intended to trick the user. \\
The final output should include the score (from the range 0-5) and your explanation for the given score. \\
The conversation for evaluation: \\
\{conversation\}
}

\subsection{Skill}
{\small \tt You are a human evaluator tasked with assessing the *skill* of the chatbot in this dialogue. \\
Skill means how well the chatbot executes the task or responds to the user's input. Consider how accurate, clear, and appropriate the responses are. \\
Give a score between 0 and 5, and provide a short explanation for your score. \\
Dialogue: \\
\{conversation\}
}

\subsection{Capability}
{\small \tt You are a human evaluator tasked with assessing the capability of responses. \\
Evaluate only capability (how effectively the chatbot fulfils user needs and achieves the purpose of the conversation). Do not assess any other dimension. Focus only on whether the chatbot meets or exceeds the user's expectations. \\
Give a score between 0-5 and a brief explanation for your score. \\
Dialogue to evaluate: \\
\{conversation\}
}

\subsection{Empathy}
{\small \tt You are an expert evaluator tasked with assessing the level of empathy of the chatbot in the conversation. Chatbot that displays high levels of empathy is the one that shows understanding, awareness, sensitivity to the feelings, thoughts, and experience of the user. \\
The final output should include the score (from the range 1-12) and your explanation for the given score. \\
The conversation for evaluation: \\
\{conversation\}
}

\subsection{Curiosity}
{\small \tt You are an expert evaluator tasked with assessing the curiosity of the chatbot in the conversation. Curiosity refers to how well the chatbot engages the user and shows interest in the responses by asking questions encouraging further interactions. \\
The final output should include the score (from the range 0-100) and your explanation for the given score. \\
The conversation for evaluation: \\
\{conversation\}
}

\subsection{Talent}
{\small \tt You are a crowdworker asked to rate the chatbot's *talent* in this conversation. \\
Talent means how naturally or intelligently the chatbot handles the conversation. \\
Was it thoughtful, clever, or showed any spark of conversational ability? Use your instinct- if it felt smart or interesting, that's talent. \\
Give a score from 0 to 5 and a short reason for your choice. \\
Dialogue: \\
\{conversation\}
}

\subsection{Overall}
{\small \tt Evaluate the following conversation between a user and a chatbot. The evaluation should be for the responses generated by the chatbot. \\
Give an integer score the scale of 0-100 to evaluate the overall impression, where 0 indicates the worst score possible and 100 indicates the best score possible. \\
The final answer must contain an integer in the range 0-100 and the reason for giving the score. \\
Here is the conversation to evaluate: \\
\{conversation\}
}

\subsection{Summarisation prompt} \label{lm-prompts-summ}
{\small \tt Prompt: \\
You are an expert copywriter tasked with shortening a chatbot's utterances from a conversation between a chatbot and a user. \\
Objective: \\
Shorten the chatbot's response while preserving its original communication style and all relevant details necessary for later evaluation. Ensure that the short version remains faithful to the chatbot's intent, tone, and structure. \\
Guidelines: \\
    - Retain all details that could be useful for evaluating the chatbot's performance. \\
    - Encode proper names that are irrelevant to the evaluation (e.g., specific phone models) using placeholders like [model-name1]. \\
    - Return the shortened dialogue as a string. \\
    - The summary must not exceed 50 words. \\
Chatbot's utterance to shorten:  \\
\{conversation\} \\
Output: A concise yet comprehensive concise version of the chatbot's response (max 50 words).
}

\section{LM Prompts examples} \label{appendix:lp-examples}
In this section we present some outputs of the distinct prompt strategies.
\subsection{Basic prompt example} \label{appendix:lm-prompts-bp}
{\small \tt Act like a human evaluator tasked with assessing the relevance of chatbot's answers. Assess only the chatbot, not the user.
The final output should include the score (from the range 0-100) and your explanation for the given score. \\
The conversation for evaluation: \\
\{conversation\}
}

\subsection{0-shot learning example} \label{appendix:lm-prompts-0sp}
{\small \tt Act like a human evaluator tasked with assessing the relevance of chatbot's answers. \\
Relevance refers to the system's ability to provide answers that are related or useful to what is happening or being talked about. \\
Please, evaluate queries of the chatbot in the following conversation by assigning it a score from the scale 0-100, where 0 means that the chatbot's answers are often irrelevant, and 100 suggests that the chatbot's answers are always relevant. \\
The final output should include the score (0-100) and your explanation for the given score. \\
The conversation for evaluation: \\
\{conversation\}
}

\subsection{1-shot learning example} \label{appendix:lm-prompts-1sp}
{\small \tt You are an expert evaluator tasked with assessing the relevance of chatbot's answers. \\
Relevance refers to the system's ability to provide answers that are related or useful to what is happening or being talked about. \\
Please, evaluate queries of the chatbot in the following conversation by assigning it a score from the scale 0-100, where 0 means that the chatbot's answers are often irrelevant, and 100 suggests that the chatbot's answers are always relevant. \\
The final output should include the score (0-100) and your explanation for the given score. \\
Here is an example excerpt of the conversation and the score this conversation received. Chatbot's and user's utterances are separated using ''';'''. \\
Excerpt from the example conversation: '''\{excerpt\}''' \\
Score for the example conversation: '''\{score\}''' \\
The conversation for evaluation: \\
\{conversation\}
}

\subsection{Few-shots learning example} \label{appendix:lm-prompts-fsp}
{\small \tt Act like a human evaluator tasked with assessing the relevance of chatbot's answers. \\
Relevance refers to the system's ability to provide answers that are related or useful to what is happening or being talked about. \\
Please, evaluate queries of the chatbot in the following conversation by assigning it a score from the scale 0-100, where 0 means that the chatbot's answers are often irrelevant, and 100 suggests that the chatbot's answers are always relevant. \\
The final output should include the score (0-100) and your explanation for the given score. \\
Here are the examples of the excerpts of the conversations and the score these conversations received. Chatbot's and user's utterances are separated using ''';'''. \\
Excerpt from the example conversation: '''\{excerpt1\}''' \\
Score for the example conversation: '''\{score1\}''' \\
Excerpt from the second example conversation: '''\{excerpt2\}''' \\
Score for the second example conversation: '''\{score2\}''' \\
Excerpt from the third example conversation: '''\{excerpt3\}''' \\
Score for the third example conversation: '''\{score3\}''' \\
The conversation for evaluation: \\
\{conversation\}
}

\subsection{Self-consistency prompting example} \label{appendix:lm-prompts-scp}
{\small \tt Act like a human evaluator tasked with assessing the relevance of chatbot's answers. Assess only the chatbot, not the user. \\
Relevance refers to the system's ability to provide answers that are related or useful to what is happening or being talked about. \\
Rate the chatbot's relevance on a scale from 0 to 100, where: \\
- 0-20: Very low relevance - The chatbot's responses are mostly irrelevant or off-topic. Users may find the answers confusing or unhelpful. \\
- 21-40: Low relevance - The chatbot provides some relevant information, but many responses are not aligned with the user's queries. Users may struggle to find useful insights. \\
- 41-60: Moderate relevance - The chatbot's answers are somewhat relevant, with a mix of useful and irrelevant information. Users may find some value but will likely encounter inconsistencies. \\
- 61-80: High relevance - The chatbot generally provides relevant and useful answers. Most responses align well with user queries, though occasional irrelevant information may still appear. \\
- 81-100: Very high relevance - The chatbot consistently delivers highly relevant and useful responses. Users can rely on the answers to be directly related to their queries, enhancing their experience significantly. \\
Return the score (0-100) along with a concise explanation of why the chatbot received that score. Think like a domain expert and check the validity of your score. Fix the score if needed. \\
Dialogue for Evaluation: \\
\{conversation\}
}